\documentclass{article}

\usepackage{microtype}
\usepackage{graphicx}
\usepackage{subfigure}
\usepackage{booktabs} 
\usepackage{hyperref}
\usepackage{cite}
\usepackage{xurl}
\usepackage{placeins}
\usepackage{balance}

\usepackage[accepted]{icml2025}

\usepackage{amsmath}
\usepackage{amssymb}
\usepackage{mathtools}
\usepackage{amsthm}

\usepackage[capitalize,noabbrev]{cleveref}

\theoremstyle{plain}

\theoremstyle{definition}

\theoremstyle{remark}

\usepackage[textsize=tiny]{todonotes}

\icmltitlerunning{Geometric Principles for Machine Learning of Dynamical Systems}
\begin{document}

\twocolumn[
\icmltitle{Geometric Principles for Machine Learning Dynamical Systems}




\begin{icmlauthorlist}
\icmlauthor{Zack Xuereb Conti}{turingZXC}
\icmlauthor{David J Wagg}{sheff}
\icmlauthor{Nick Pepper}{turingNP}
\end{icmlauthorlist}

\icmlaffiliation{turingZXC}{Data-centric engineering, The Alan Turing Institute, London, UK}
\icmlaffiliation{sheff}{School of Mechanical, Aerospace and Civil Engineering, University of Sheffield, Sheffield, UK}
\icmlaffiliation{turingNP}{Project Bluebird, The Alan Turing Institute, London, UK}

\icmlcorrespondingauthor{Zack Xuereb Conti}{zxuerebconti@turing.ac.uk}

\icmlkeywords{Machine Learning, ICML}

\vskip 0.3in
]




\begin{abstract}

Mathematical descriptions of dynamical systems are deeply rooted in topological spaces defined by non-Euclidean geometry. This paper proposes leveraging structure-rich geometric spaces for machine learning to achieve structural generalization when modeling physical systems from data, in contrast to embedding physics bias within model-free architectures. We consider model generalization to be a function of symmetry, invariance and uniqueness, defined as a topological mapping from state space dynamics to the parameter space. We illustrate this view through the machine learning of linear time-invariant dynamical systems, whose dynamics reside on the symmetric positive definite manifold.





\end{abstract}

\section{Introduction} 
\label{intro}

Geometry underpins the foundations of physics so fundamentally that Newton did not use equations to present his theory; instead he presented it geometrically and proved theorems in the Principia using techniques of Euclidean geometry \cite{maudlin2012}. The subject matter–motion in space– is itself geometrical. Similarly, Kepler believed the structure of the cosmos was not rooted in numbers but in geometry \cite{sym13112100}. 


Most real-world physical systems are governed by underlying dynamical systems. This implies that when modeling physical systems, whether from physics or data-driven, we can reasonably assume that their underlying behavior is driven by lower dimensional structures despite their potentially high-dimensional representation. Such systems exhibit an inherent structure that governs their behavior, making them amenable to mathematical description and generalization across varying conditions. From classical mechanics we know that such structures can be fundamentally described geometrically \cite{abraham1994foundations}.

In contrast, the recent rise of `model-free' data-driven methodologies for modeling physical systems, such as deep learning algorithms,  challenges this structure-grounded notion of generalization. Model-free approaches are widely popular for their ability to capture complex patterns from data with little to no prior knowledge of the underlying physical mechanics. However, it has been well documented \cite{Wang2020, Kutz2022} that deep learning models for modeling dynamical systems frequently fail to generalize when exposed to new initial conditions or parameter regimes, often necessitating retraining. Their opaque architecture conditions their generalization success entirely to the representation of the phenomenon in the data. Specifically, when training a neurual network for example, the algorithm itself (stochastic gradient descent) is implicitly regularizing the solution thus, conditioning generalization of the trained model success entirely to  the quality of the data \cite{zhang2017}. Measuring their generalization is often reduced to the `fit' via averaged local point-based techniques which in turn provides limited insight into the global behavior of the system under varying initial conditions and parameters. These points highlight a fundamental limitation in their ability to capture the governing structures that underlie physical systems, reliably and parsimoniously \cite{Kutz2022}. 



In response to these caveats, the emergence of physics-based machine learning aims to reduce the dependency on representation in data by encoding available scientific knowledge about the phenomena in the form of biasing strategies. 

\subsection{Physics-based Machine Learning}




More specifically, \citeauthor{karniadakis2021physics} (\citeyear{karniadakis2021physics}) suggested the following categories for biasing strategies: \textit{observational bias}, \textit{inductive bias}, and \textit{learning bias}, each with distinct methodologies. Observational bias relies on high-quality datasets, such as using neural operators to approximate PDE solutions from comprehensive training data~\cite{kovachki2023neural}. Inductive bias embeds physical constraints directly into model architectures, like equivariant neural networks that enforce symmetry invariances or PDE-constrained activation functions~\cite{ranftl2022connection}. Learning bias incorporates physics-informed constraints into loss functions during training, as seen in physics-informed neural networks (PINNs), which enforce governing equations while fusing observational data~\cite{raissi2020hidden}. An extensive review of each of these categories can be found in \citet{muther2022}, \citet{wang2021}, \citet{willard2020integrating}, \citet{hao2022physics}.


However, while incorporating physics as biases into deep learning architectures can contribute towards scientific discoveries of physical complexities that were not previously captured, the caveats discussed above persist and hamper the ability of these methods to generalize. We argue that extensive knowledge of structure-rich spaces from classical mechanics and differential geometry remains an underutilized opportunity for developing data-driven models that can generalize at a structural level rather than merely fitting data with predefined constraints. 

Related emerging topic areas and contributions at this intersection include geometric deep learning \cite{bronstein2021geometricdeeplearninggrids}, geometry-informed machine learning  \cite{Hern_ndez_2023, champion2019data, Floryan2022}, and thermodynamics-informed deep learning \cite{BARBARESCO2022107} where structural assumptions about underlying phenomena play a central role to modeling physical systems from data using machine learning and deep learning techniques. 

In this paper we suggest leveraging physical understanding of phenomena when modeling physical systems by creatively constraining  machine learning within naturally generalized, structure-rich representations of dynamical systems. The latter is a well-established topic in classical mechanics where such representations are rooted in non-Euclidean geometric spaces and governed by principles such as symmetry, invariance and uniqueness. 

\subsection{Paper Layout}

In \cref{case-eg} we introduce a case example of a linear dynamical system whose structural generalization is discussed from first principles in \cref{generalization}, as a function of its symmetric and invariant structural composition. Subsequently, \cref{sec:ml} illustrates how the geometric representation of the demonstrative dynamical system could be leveraged to preserve generalization by nudging towards target measurement data. The results from a benchmark study are then discussed in  \cref{sec:results}. 

\section{Case Example}
\label{case-eg}

For illustrative purposes, we assume a simple material system undergoing one-dimensional heat transfer laterally along its thickness. The heat transfer conduction dynamics can be described by the general one-dimensional heat conduction equation given by:



\begin{equation}
\frac{\partial u(x,t)}{\partial t} = \frac{k}{\rho c} \frac{\partial^2 u(x,t)}{\partial x^2} + \frac{q(x,t)}{\rho c},
\label{eqn:heatpde}
\end{equation}
where: $u(x,t)$ is the thermal conductivity of the material ($W/mK$), $\rho$ is the density of the material ($kg/m^3$), $c$ is the specific heat capacity of the material ($J/kgK$) and $q(x,t)$ is the internal heat generation per unit volume ($W/m^3$).



\subsection{Discretization}

To model the dynamics of a physical material, we must approximate the continuous temperature field in \cref{eqn:heatpde} by reducing it to a discrete set of nodes \( j = 1, 2, \dots, N \), with each node representing a lumped thermal mass \( C_j \) connected by thermal resistances . This is commonly referred to as a lumped-parameter model: 

\begin{equation}
C_j \frac{dT_j}{dt} = \sum_k \frac{1}{R_k} (T_k - T_j) + Q_j,
\label{eqn:lumped}
\end{equation}



where $R_k = \frac{\Delta x}{ka}$ is the thermal resistance between nodes $j$ and $k$, $\Delta x$ is the distance between nodes, $a$ is the cross-sectional area and $k$ is the thermal conductivity. The heat balance is composed of: the temporal derivative of temperature at each node \cref{eqn:temporal}, the spatial derivative between nodes \cref{eqn:spatial}, and the heat stored at each node \cref{eqn:node}. \cref{eqn:spatial} is approximated via the finite difference method via:
\begin{align}
    \label{eqn:temporal}
    \frac{\partial u(x,t)}{\partial t} &\approx \frac{dT_j}{dt}, \\
    \label{eqn:spatial}
    \frac{\partial^2 u(x,t)}{\partial x^2} &\approx \frac{1}{R_k} (T_k - T_j),\\
    \label{eqn:node}
    C_j \frac{dT_j}{dt}& = \text{Net heat flow into the node}.
\end{align}

For illustrative purposes, we limit our discrete approximation of the physical material layer to two temperature states $T_{ext1}, T_{ext2} \in \mathbf{T}$ (on either side of the material thickness) where $T_{ext1}$ only is directly influenced by an external forcing $T_{ext}$ which is the ambient temperature influencing the dynamics through convection. Thus, from \cref{eqn:lumped} we arrive to the following system of ordinary differential equations (ODE): 

\begin{equation}
    \begin{split}
     & \frac{dT_{ext1}}{dt} = \frac{1}{C_{ext1}} \left[ \frac{1}{R_{\text{ext2\_ext1}}} \left( T_{\text{ext1}} - T_{\text{ext2}} \right) \right], \\
     & \frac{dT_{ext2}}{dt} = \frac{1}{C_{ext2}} \left[ 
     \frac{1}{R_{\text{ext2\_ext1}}} \left( T_{\text{ext1}} - T_{\text{ext2}} \right) \right] \\
     & + \frac{1}{R_{\text{ext\_ext2}}} \left( T_{\text{ext2}} - T_{\text{ext}} \right). \\
    \end{split}
\label{eqn:odes}
\end{equation}

\subsection{State Space Matrix Representation}
We can represent the system of ODEs in \cref{eqn:odes} as a compact linear time-invariant state space matrix representation to gain qualitative sense of the underlying structure governing the dynamical system. $\mathbf{T} \in \mathbb{R}^n$ is a state vector, $\mathbf{U} \in \mathbb{R}^{m \times 1}$ is input vector, $A \in \mathbb{R}^{n \times n}$ contains the information about the unforced dynamics of all system states $\mathbf{T}$ while $B \in \mathbb{R}^{n\times m}$ determines how the input matrix $\mathbf{U}$ (forcing) influences the states $\mathbf{T}$. The structure of $A$ is invariant to the order of the system provided the system retains the same physical interaction principles and topological connectivity between the nodes or states:


\begin{equation} 
    \label{eqn:ssm_}
    \frac{d\mathbf{T}}{dt} = A \mathbf{T} + B \mathbf{U},
\end{equation} 

\begin{equation} 
\label{eqn:ssm}
    \begin{split}
    \frac{d\mathbf{T}}{dt} & = 
         \renewcommand\arraystretch{1}
         \begin{bmatrix}
        \dfrac{-U_{ext1,ext2}}{C_{ext1}} & \dfrac{U_{ext1,ext2}}{C_{ext1}} \\
        \dfrac{U_{ext1,ext2}}{C_{ext2}} & \dfrac{-U_{ext1,ext2}-U_{ext2,ext}}{C_{ext2}}\\
        \end{bmatrix}
        \begin{bmatrix}
        T_{ext1} \\
        T_{ext2}\\
        \end{bmatrix} \\ 
        & +
        \begin{bmatrix}
        0  \\
        \dfrac{U_{ext2,ext}}{C_{ext2}} \\
        \end{bmatrix}
        \left[ T_{ext} \right]. \\     
    \end{split}
\end{equation} 

where $U$ is the thermal transmittance ($W/m^2K$) and is equivalent to $\frac{1}{R}$. In order to solve the system in \cref{eqn:ssm}, we must convert time-continuous $A$ to discrete-time $\Phi_A$ via the matrix exponential expansion, as follows: 

\begin{equation} \label{Phi_A} 
\Phi_A = e^{At} \text{ and }  \Phi_B=A^{-1}(e^{At}-I)B
\end{equation} 

where $t$ is the time-step for discretization and where:

\begin{equation} \label{Phi_A}
e^{At} = \mathbf{I}+ At + \frac{A^2t^2}{2!} + \frac{A^3t^3}{3!}
+ ... + \frac{A^kt^k}{k!} + ...,
\end{equation} 
which leads to the following equation for the discrete time dynamics:
\begin{equation} \label{eq:4}
\dot{\mathbf{T}}(t) = e^{At}\mathbf{T}(t_0) + A^{-1}(e^{At}-I)B\mathbf{U}(t_0).
\end{equation}







\section{Generalization From First Principles}
\label{generalization}

\label{final author}

The demand for generalization and parsimony are crucial attributes when modeling physical systems across domains, as models are often expected to make accurate predictions for a range of operating conditions, some of which might be underrepresented in the training data. In dynamical systems this implies achieving generalization for varying initial conditions.




We view generalization from first principles, where natural generalization of dynamical systems derived from physical theories and laws, when expressed as a system of differential equations for example, is as function of their underlying structure which can be described geometrically. 


\subsection{Structural Symmetry}

A physical system is considered symmetric if it remains unchanged under specific transformations. They key concept underlying fundamental equations of physics is conservation, as formalized in the ``conservation laws". These laws are deeply tied to the principle of symmetry, which states that the laws of nature remain invariant under certain transformations \cite{poincare1902}. From classical mechanics, we know that such structures are expressed in terms of abstract invariant geometric objects. \citeauthor{north2009structure} (\citeyear{north2009structure}) argues that physics adheres to the methodological principle that the symmetries in the laws match the symmetries in the structure of the world. Thus, in this paper we argue that if when approximating physical systems through discretization techniques for example, such governing symmetries present in the laws must be preserved, to preserve structural generalization.

In our discretization of \cref{eqn:heatpde}, we can observe how the symmetry governing the conduction phenomenon is preserved in the symmetry of $A$ in \cref{eqn:ssm}, as follows:


\begin{equation}
    \begin{split}
        A & =  -C^{-1} G \\
          & = -
        \begin{bmatrix}
        \dfrac{1}{C_1} & 0 \\
        0 & \dfrac{1}{C_2}
        \end{bmatrix} 
        \begin{bmatrix}
        \dfrac{1}{R_1} + \dfrac{1}{R_{12}} & -\dfrac{1}{R_{12}} \\
        -\dfrac{1}{R_{12}} & \dfrac{1}{R_2} + \dfrac{1}{R_{12}}
        \end{bmatrix} \\
    \end{split}
\end{equation}

such that,

\begin{equation}
    A = A^T \longrightarrow{\lambda_i} \in \mathbb{R}.
\end{equation}


The symmetry of the conductance matrix $G$ directly imposes real eigenvalues and orthogonal eigenvectors. The structure of system matrix $A$ which contains the state dynamics is governed by the eigenstructure of $A$ by the following decomposition: 

\begin{equation}
\label{eqn:eigdecomp}
    A = \mathbf{V} \mathbf{\Lambda} \mathbf{V}^{-1},
\end{equation}

where $(\mathbf{\Lambda} = \text{diag}(\lambda_1, \lambda_2, \dots, \lambda_N)$ is the diagonal matrix of eigenvalues and $\mathbf{V}$ is the matrix of eigenvectors of $\mathbf{A}$. 

\begin{figure}[ht]
\vskip 0.2in
\begin{center}
\centerline{\includegraphics[width=\columnwidth]{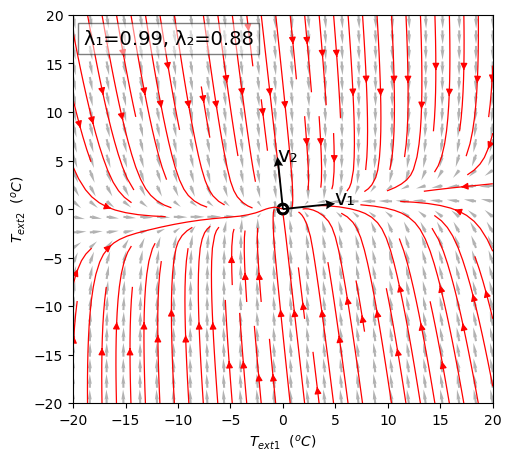}}
\caption{{Phase portrait of the 1D heat conduction dynamics governed by the eigenstructure of $A$.}}
\label{fig:phaseportrait}
\end{center}
\vskip -0.2in
\end{figure}

The generalizability of the matrix to describe global dynamics i.e. for all initial conditions, is governed by the symmetry of the matrix, which in turn reflects the symmetry of the resistances and conservative dynamics. The eigenvalues $\mathbf{\Lambda}$ and eigenvectors $\mathbf{V}$ determine the trajectory behavior at all initial conditions in the phase space, as illustrated by the phase portrait of the unforced dynamics (only considering \cref{Phi_A}) illustrated in \cref{fig:phaseportrait}. The trajectories decay along the eigenvectors towards the stable node. 

When $A$ is symmetric, its eigenvalues are guaranteed to be real and its eigenvectors orthogonal (or can be orthonormalized). Symmetric matrices are often associated with conservative dynamics or systems without directional biases (e.g. no preferential heat flow or non-reciprocal effects). Symmetry ensures predictable and stable dynamics since the eigenbasis is `well-behaved' and the modes evolve independently (via orthogonality). On the other hand, if  $A$ is asymmetric arising from non-reciprocal resistances or external forcing, the eigenvalues may be complex in which case oscillatory modes are present and transient dynamics can dominate the trajectories before reaching steady state. 

\subsection{Structural Invariance}

There can be different mathematical formulations of a given physical theory defined by different coordinate systems describing the space, such as Lagrangian and Hamiltonian versions of classical mechanics. However, the underlying structure of the mathematical space is described by coordinate-independent objects whose intrinsic nature is invariant to change when varying the description of the mathematical space, i.e. the coordinate system, or even the physical parameters of the system. In other words, there is a structure to these objects \cite{north2009structure}.

\subsubsection{Coordinate space}

Let us isolate/decouple the state space dynamics component $A\mathbf{T}$ from from the forcing component $B\mathbf{U}$ in \cref{eqn:ssm_} and introduce a new coordinate system via an invertible transformation matrix, $P$, where $\boldsymbol{y}=P{\pmb T}$ such that now:


\begin{equation} \label{eqn:coordT}
    \frac{d\mathbf{y}}{dt} = P \frac{d\mathbf{T}}{dt}
    \leadsto  \frac{d\mathbf{y}}{dt} = PAP^{-1}\mathbf{y}.
\end{equation} 

The new system matrix $A'$ in the transformed coordinate is:

\begin{equation}
 A'=PAP^{-1}.
\end{equation} 

The symmetric structure of $A$ does not change under coordinate transformation $P$. This is evidenced by observing the eigenstructure where the eigenvalues of $A$ and $A'$ are identical by their roots: 

\begin{equation}
 \det(A-\lambda I) = \det(PAP^{-1} - \lambda I)=\det(A'-\lambda I). 
 \end{equation} 

Similarly for the eigenvectors of $A$ where, if $v \in \mathbf{V}$ is an eigenvector corresponding to an eigenvalue $\lambda \in \mathbf{\Lambda}$, then $Pv$ is an eigenvector of $A'$ corresponding to the same eigenvalue $\lambda$, as follows: 

\begin{equation}
    Av = \lambda v \Longrightarrow PAP^{-1}(Pv)=\lambda(Pv).
 \end{equation} 

Thus, when transforming the eigendecomposition in \cref{eqn:eigdecomp}, $\mathbf{\Lambda}$ and orthogonality in $\mathbf{V}$ is preserved, as follows: 

\begin{equation}
    A' = (P\mathbf{V}) \mathbf{\Lambda} (P\mathbf{V})^{-1}
 \end{equation} 
  
This invariance is reflected in the phase portrait which remains qualitatively the same under a similarity transformation where trajectories are simply re-parameterized in the new coordinate system. The structural reason behind this invariance, belongs to the fact that both $A$ and $A'$ related by $A'=PAP^{-1}$ lie on the same equivalence class on a manifold of matrices. 


\subsubsection{Parameter space}

Similarly, the invariance of the structure underlying a physical system extends to the parameter space where perturbations in system parameter values do not modify the structure of the underlying dynamics. This is because the dynamics reside on a manifold $\mathbf{M}$ of structurally equivalent systems, where variations in the parameter space map directly to correspond to moving along a paths or trajectories within this manifold.

\begin{figure}[ht]
    \vskip 0.2in
    \begin{center}
    \centerline{\includegraphics[width=\columnwidth]{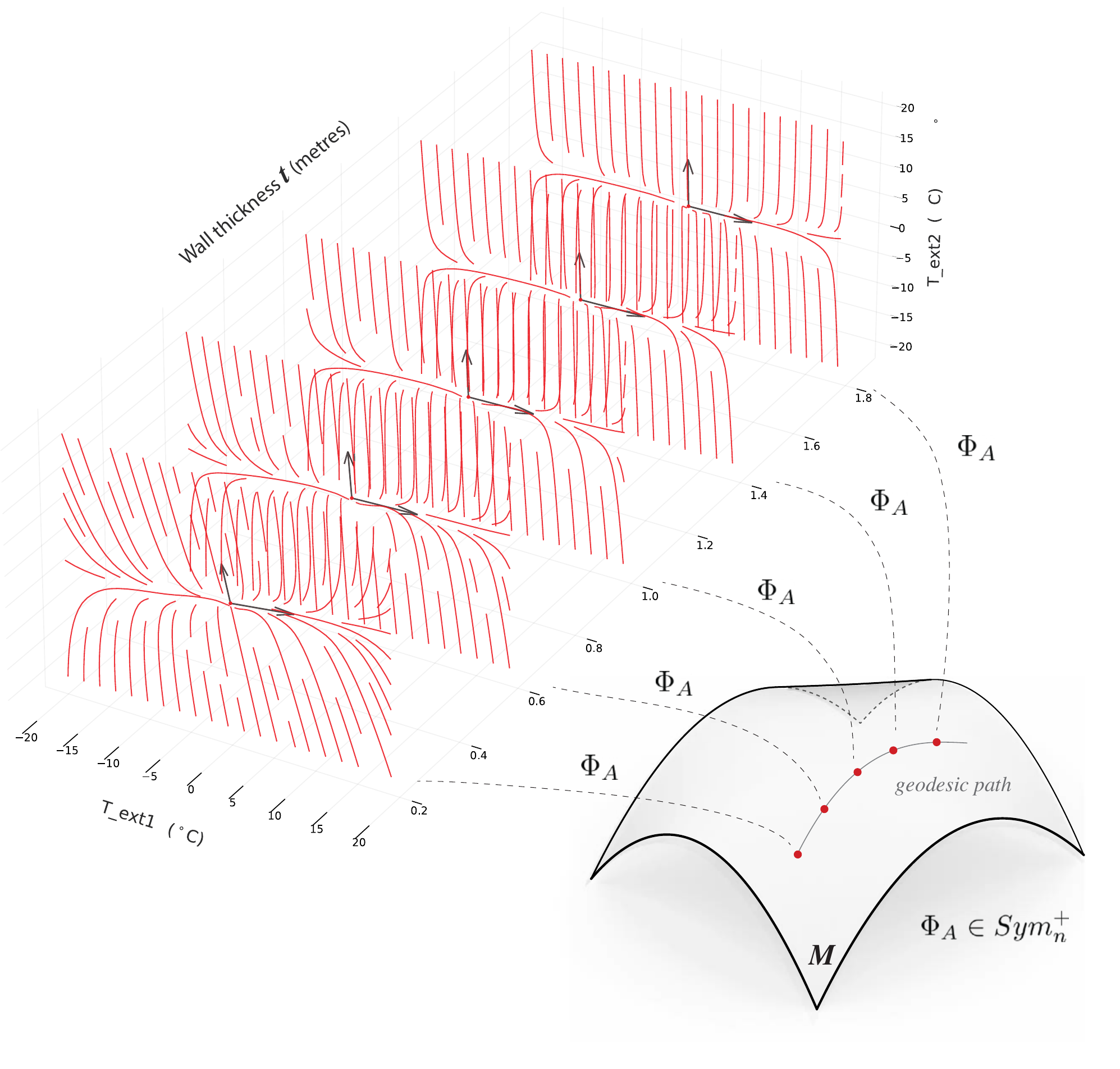}}
    \caption{Perturbing parameters of a dynamical system traces a geodesic path on the manifold. In this example, varying the material thickness of our case example results in smooth deformations of the phase portrait geometry as a function of a geodesic path on the $Sym_n^+$ matrix manifold.}
    \label{fig:allportraits}
    \end{center}
    \vskip -0.2in
\end{figure}

So for example, varying the heat transfer coefficient $U$ in \cref{eqn:ssm} as a function of material thickness $t$ or material conductance $k$, directly leads to smooth geometric deformations of the phase portrait geometry as illustrated in Figure~{\ref{fig:allportraits}}. This can explained geometrically as a scalar transformation of the eigenvalues coupled with a rotational transformation of the eigenvectors, while preserving their orthogonality and steady state point of equilibrium. 


\subsubsection{Manifold theory}

We can think of the mathematical descriptions of dynamical systems as belonging to an underlying hierarchical structure of increasingly generalized spaces, where perturbations in a system's parameters are reflected as paths or trajectories on a manifold. Within this hierarchy, the behavior of a system at any given point can be characterized using a well-established set of operators. Governing laws used to explain the temporal behavior of known physical phenomena are often `smooth'. A smooth manifold is a topological space that locally resembles Euclidean space and is equipped with a globally-defined smooth structure. This means smooth functions can be defined as differentiable maps and tangent spaces on the manifold. Further reading can be found in \citeauthor{asselmeyer2007exotic} (\citeyear{asselmeyer2007exotic}).

As an example, the system matrix $A$ in  \cref{eqn:ssm} belongs to the symmetry matrix manifold $Sym_n$ where $A=A^T$ and which is a Euclidean (flat) subspace of the space of all matrices $\mathbb{R}^{n \times n}$. Its time-discretization $\Phi_A$ belongs to the symmetric positive definite (SPD) manifold $Sym_n^+$ which is a non-Euclidean space (curved) and a submanifold of $Sym_n$ where matrices are symmetric but specifically, positive definite. For a matrix to be positive definite, its eigenvalues must be positive (i.e. $Re(\lambda_i) > 0)$). The SPD manifold $\mathcal{M}$ is a smooth differentiable topological space equipped with an invariant Riemannian structure (i.e. Riemannian manifold). The structure facilitates a Riemannian metric that varies smoothly from point to point where every point is equivalent to a unique and valid physical system. For further reading on Riemannian metrics, see \citeauthor{sommer:hal-02341901} (\citeyear{sommer:hal-02341901}). For each system matrix $\Phi_A \in \mathcal{M}$, it is possible to compute a tangent space $\mathcal{T}_{\Phi_A}\mathcal{M}$. The operator used to map from a point on the manifold to its tangent space is given by the logarithmic map $\mathrm{Log}_{\Phi_A} (m):\mathcal{M} \rightarrow \mathcal{T}_{\Phi_A}\mathcal{M}$ while the inverse is given by the exponential map $\mathrm{Exp}_{\Phi_A} (m):\mathcal{T}_{\Phi_A}\mathcal{M}\rightarrow \mathcal{M}$. Therefore, we can interpret the tangent space at a given system $\Phi_A$ in \cref{eqn:ssm} on the $Sym_n^+$ manifold $\mathcal{M}$ as the linearized space of all possible infinitesimal perturbations of $\Phi_A$ that preserve the symmetric structure. 

Thus, when interpreted geometrically, the latter is equivalent to \cref{Phi_A} implying that time-discretization of $A$ is a projection from continuous-time dynamics residing in the Euclidean space of all possible symmetric dynamical systems $Sym_n$ to the non-Euclidean space $Sym_n^+$ of symmetric but stable discrete-time dynamical systems, by means of their positive eigenvalues implying positive definiteness. In further detail, $e^{At}$ is a bilinear map that geometrically maps the complex $\textit{s}$-plane to the complex unit circle in the $\textit{z}$-plane where system stability is preserved by wrapping the stable eigenvalues located in the left half-plane (i.e., $Re(\lambda_i) < 0)$) onto the unit circle in the $\textit{s}$-plane, implying positive-definiteness. System matrices $\Phi_A$ that lie on the surface of the SPD manifold are positive \textit{semi}-definite attributed to their low-rank are said to be \textit{bistable} due to some eigenvalues $Re(\lambda_i)\geq 0)$. In general, as you move towards the boundary in the stratified space composing the SPD manifold, the matrix loses rank, meaning that fewer independent eigendirections remain for the system trajectories to evolve in. For further reading on the role of symmetry in dynamical systems, see \citeauthor{marsden2013introduction} \citeyear{marsden2013introduction}.

In the presented example, the structural symmetry and invariance of the Riemannian manifold affords the natural generalizability across parameter perturbations when represented using structure-rich representations.

\section{Leveraging Geometry}
\label{sec:ml}

Having illustrated the geometric connection between time evolving state trajectories of a dynamical system in the phase space and structure-preserving parameter perturbations within its underlying manifold, it becomes natural to leverage this representation to machine learn global dynamics from measured data. 

In what follows, the heat transfer example from \cref{case-eg} is used to demonstrate how a discretized representation of the system may be learned by `nudging' the discretized state space model in \cref{eqn:ssm} closer towards a dynamical system that represents the stable dynamics underlying the measured temperature data. At the same time, the symmetric and positive definite structure governing the natural generalization and stability of the heat equation in \cref{eqn:heatpde} is preserved.



Our goal is for the resulting model to generalize beyond the initial condition it was trained on, moving away from black-box approaches. Specifically, we aim to uncover the underlying eigenstructure that governs the measured system's behavior, by perturbing or `nudging' the physics-approximated state space model within the underling $SPD$ manifold. In the control/dynamical systems community, this could be interpreted as manifold-constrained system identification (SID).  



\subsection{Data}
A sequence of one year's worth of synthetic, hourly measurement temperature data (8,759 hours) $\{\mathbf{T_1},\dots, \mathbf{T_n}\}$ was generated via a high-fidelity numerical analysis of a homogeneous material system using EnergyPlus, which is a simulation software adopted widely in the thermal energy modeling community \cite{energyPlus}. The selected physical and thermodynamic properties are found in \cref{table:simulation}, while the ambient dry bulb temperature acting as a forcing ${\mathbf{U_1}, \dots, \mathbf{U_n}}$ was obtained from a historical weather file located in London \cite{ladybugweather}, and used as an input for the numerical simulation.





\subsection{Structure-informed Optimization}


We start with an initial state space model $A$ that is derived from Physics but misspecified (see \cref{table:simulation}), and which is used as an initial guess at the start the optimization. Sensor data is received at discrete time-steps, hence it is necessary to reformulate \cref{eqn:ssm} into its discrete-time form via \cref{Phi_A}, as follows: 
\begin{align}
    \mathbf{T}_{t+1}=\Phi_A\mathbf{T}_t+\Phi_B\mathbf{U}_t.
\end{align}
The optimization goal is to learn a new state space model whose matrices denoted $\hat{\Phi}_A$ and $\hat{\Phi}_B$ better fit the target measurement data. The matrices are parameterized as tensors of size ${n \times n}$ and ${n \times m}$, respectively. The optimization problem described above may be stated as: 

\begin{align}
   \hat{\Phi}_A, \hat{\Phi}_B&=\underset{\Phi_A, \Phi_B}{\text{arg min}}\;\mathcal{J}(X|\Phi_A, \Phi_B),\\
   &\text{s.t.}\, \Phi_A^\top=\Phi_A\, \text{and}\, \mathbf{T}^\top\Phi_A \mathbf{T}>0 \,\{\mathbf{T}|\mathbf{T}\in\mathbb{R}^2\},
   \label{eq:opt}
\end{align}
where the loss function, $\mathcal{J}$, is defined as:
\begin{align}
    \mathcal{J}(X|\Phi_A, \Phi_B)=\sum^{n-1}_{i=1} \|\Phi_A\mathbf{T}_{i}+\Phi_B\mathbf{T}_i-\mathbf{T}_{i+1}\big\|^2_2.
\end{align}

We distinguish between the optimization for $\hat{\Phi}_A$ and for $\hat{\Phi}_B$. To preserve the symmetric positive structure, we adopt the Riemannian Adaptive Optimization Method (RAdam) \cite{bécigneul2019} to estimate the $\hat{\Phi}_A$ tensor where gradient updates follow the curved geodesic. The Riemannian gradient is given by: 

\begin{equation}
    \nabla_{\Phi_A} \mathcal{J}(\Phi_A^{(i)}),
\end{equation}

and therefore, gradient updates follow the curved geodesic by projecting the gradient onto the tangent space as follows: 

\begin{equation}
    \Phi_A^{(i+1)} = \text{exp}_{\Phi_A^{(i)}} \left( -\eta . \dfrac{\hat{m_i}}{\sqrt{\hat{v_i}}}\right),
\end{equation}

where where $\eta$ is a user determined learning rate, $\hat{m_i}$ and $\hat{v}_i$ are the bias-corrected first and second moment estimates, respectively. The RAdam was implemented using $\texttt{geoopt}$ Python library \cite{geoopt}. On the other hand, gradient updates for learning $\hat{\Phi}_B$ were computed in flat Euclidean space using Adam \cite {kingma2017} in $\texttt{torch.optim}$ Python library \cite{NEURIPS2019_9015}. 

In an alternative approach, $\hat{\Phi}_A$, may also be parameterized by the lower Cholesky decomposition via $\hat{\Phi}_A = LL^T$ to ensure optimization stays within the SPD manifold. The non-zero elements of $L$ could be found through an optimization in Euclidean space.  

\section{Results}\label{sec:results}

\begin{table*}
\begin{center}

\begin{tabular}{lllll}
\toprule
Method & \multicolumn{2}{c}{$T_{ext1}$} & \multicolumn{2}{c}{$T_{ext2}$}  \\
 & London & Chicago & London & Chicago \\
\midrule
LSSM from Physics ($\Phi_A, \Phi_B$) & 2.86e+00 & 1.07e+01 & 6.06e-01 & 2.10e+00 \\
Rie opt ($\hat{\Phi}_A, \hat{\Phi}_B$) & \textbf{4.00e-01} & \textbf{1.36e+00} & 5.07e-01 & \textbf{1.79e+00} \\
Euc opt ($\hat{\Phi}_A, \hat{\Phi}_B$) & 1.28e+00 & 3.35e+00 & 5.80e-01 & 1.98e+00 \\
RF & 6.81e-01 & 2.41e+01 & 2.32e-01 & 1.63e+01 \\
XGBoost & 5.02e-01 & 2.23e+01 & \textbf{1.06e-01} & 1.33e+01 \\
LSTM & 2.57e+01 & 4.01e+01 & 6.10e+00 & 7.85e+00 \\
\bottomrule
\end{tabular}
\caption{MSE error of models applied to test datasets.}
\label{table:mseloss}
\end{center}
\end{table*}

The fundamental difference between a structure-aware machine approach and using model-free approaches is that the learning acts structurally not locally at each data point or window of data points. This is evidenced by the fact that we are learning the vector field governing the phase space as illustrated in \cref{fig:london}, bottom left. To highlight this, we repeated the same modeling task across three popular model-free time-series modeling approaches namely: Random Forest ($\mathbf{RF}$), eXtreme Gradient Boosting ($\mathbf{XGBoost}$), and Long Short-Term Memory networks ($\mathbf{LSTM}$). Additionally, we repeat the system identification of the linear state space model where instead, $\hat{\Phi}_A$ and $\hat{\Phi}_B$ tensor elements are learned using only Euclidean gradient updates - denoted as $\mathbf{Euc \text{ } opt}$.

Our measurable goal in this exercise is to evaluate the generalization of the learned dynamics when predicting $T_{ext1}$ and $T_{ext2}$ accurately for: a) future time-steps and b) varying initial conditions. 


\subsection{Generalizing in-time}
From \cref{fig:london} we observe how all considered modeling approaches seem to capture the trend from a statistical fit perspective, evidenced by the MSE loss of their prediction comparatives for the unseen time-steps in \cref{table:mseloss}. However, on comparing the training convergence of the model-free MSE loss/epoch in \cref{fig:rf_sweep} ($\mathbf{RF}$, $\mathbf{XGBoost}$) and \cref{fig:lstm_train} ($\mathbf{LSTM}$) versus the model-based $\mathbf{Euc \text{ } opt}$ and $\mathbf{Rie \text{ } opt}$ in \cref{fig:lssm_train} we note a significantly quicker convergence over the model-free approaches while also noting the smoothness of the convergence attributed to the uniqueness of the solutions on the smooth SPD manifold endowed with the Riemannian metric. On further comparison, $\mathbf{Rie \text{ } opt}$ converges quicker than $\mathbf{Euc \text{ } opt}$ due to the SPD-bound gradients, see Appendix~\ref{sec:app_lssm}.  

\subsection{Generalizing for varying initial conditions}

\begin{figure*}[ht]
\vskip 0.2in
\begin{center}
\centerline{\includegraphics[width=1.87\columnwidth]{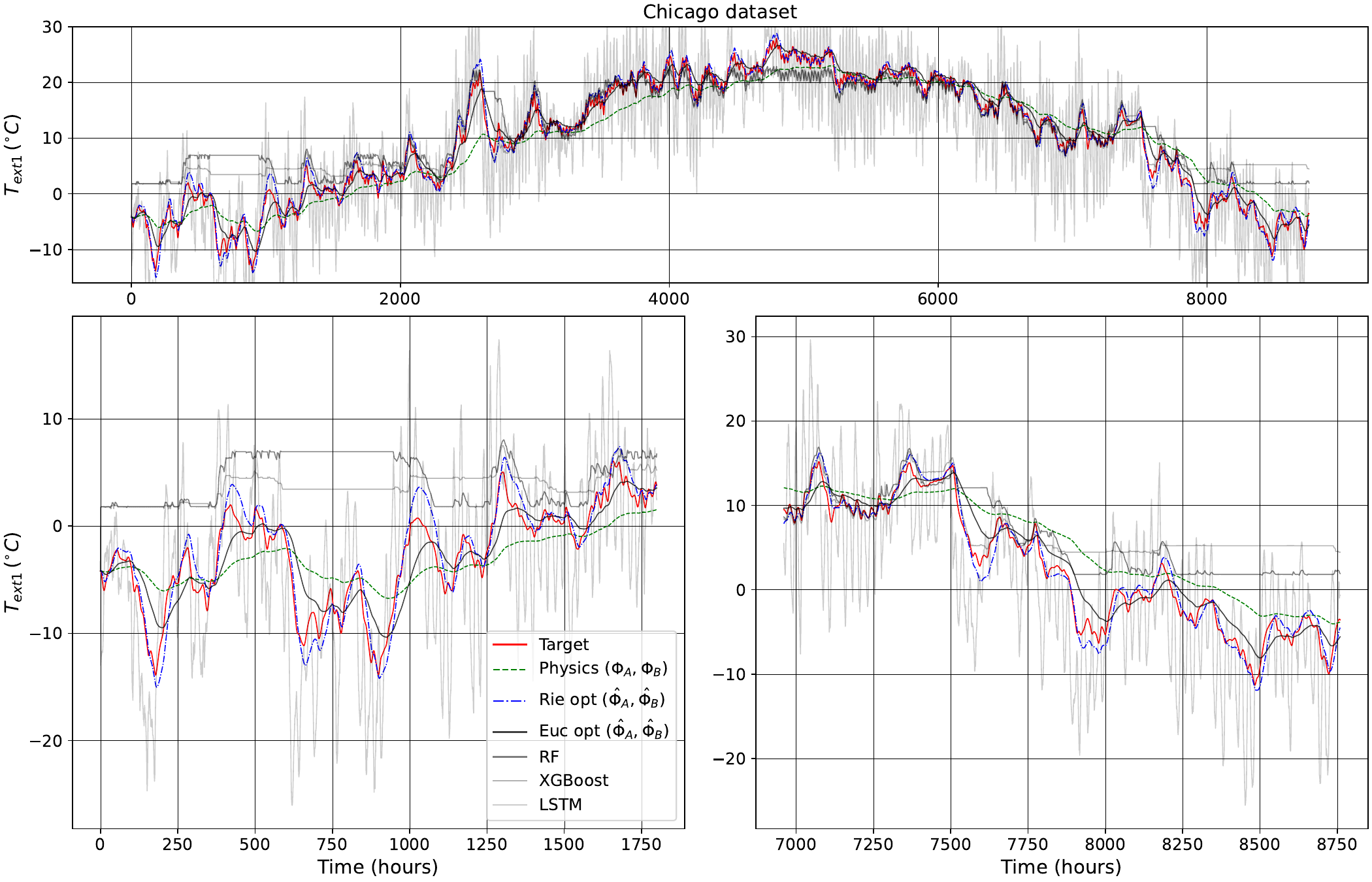}}
\caption{Model testing on the unseen Chicago dataset.}
\label{fig:chicago}
\end{center}
\vskip -0.2in
\end{figure*}

Point-wise validation metrics such as the MSE loss do not provide us with any insight into whether the model-free approaches have captured the structure of the underlying Physics and thus, we cannot gain insight into how they will generalize when exposed to unseen perturbations of the initial condition. To test such generalization, we expose all models to an unseen sequence of hourly forcing temperatures $T_{ext}$ across one year, located in Chicago \cite{ladybugweather} where temperatures exhibit different seasonal extremes to London (\cref{fig:chicago}). 

The underlying thermal dynamics of the material system is invariant of the forcing, implying that if a model has successfully captured the relationship between the forcing and the unforced dynamics, it should generalize for the unseen initial condition. 

\begin{figure*}[ht] 
\vskip 0.2in
\begin{center}
\centerline{\includegraphics[width=1.87\columnwidth]{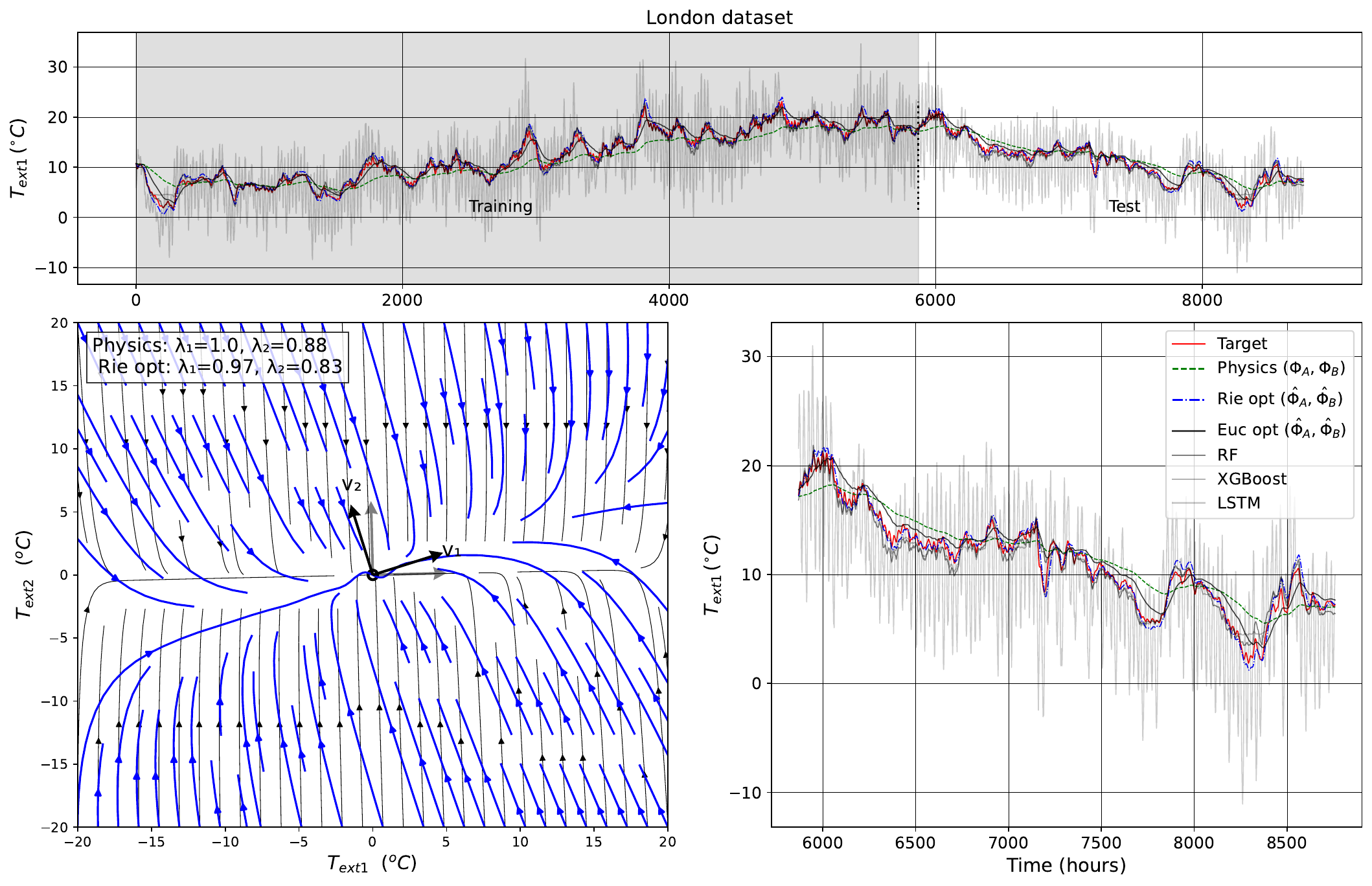}}
\caption{Model training and testing on the London dataset.}
\label{fig:london}
\end{center}
\vskip -0.2in
\end{figure*}

Observing both the MSE loss in \cref{table:mseloss} and the time-series fit when predicting the indirectly forced state $T_ext1$ in \cref{fig:london} we can instantly note how all model-free approaches demonstrate instability in contrast with, the model-based approaches $\mathbf{Rie \text{ } opt}$ and $\mathbf{Euc \text{ } opt}$ which demonstrate global stability in capturing the dynamics, in particular the former, as illustrated by the nudged phase portrait in \cref{fig:london} (bottom, left). This is likely due to the structure-aware approach learning the decoupled dynamics, as opposed to the investigated model free approaches, that learn the forced response of the system as a time series.


\subsection{Discussion}

The uniqueness of system matrices on the $SPD$ manifold regularizes the machine learning algorithm to learn globally valid system models, prevent learning physically invalid or locally incontinent representations which might arise if the system matrix $\Phi_A$ were allowed to deviate arbitrarily, and ensure generalization across parameter variations while preserving the underlying structure of dynamics. 

We recommend ensuring the preservation of the geometric features such as symmetry when selecting the spatial discretization of the PDE. Leveraging geometric features in this way could improve the ability of ML models to capture global decoupled dynamics and hence improve the generalization of ML models to unseen operating conditions.  

Enforcing structural preservation in machine learning could be further improved parsimoniously through Lie Group parametrization of the dynamical system where the matrix exponential map in \cref{Phi_A}  is interpreted as a map from the Lie algebra (tangent space) to the Lie group on the SPD manifold. This will be presented in future work.  




\section*{Impact Statement}


Achieving stable generalization and physical interpretability are crucial criteria for the reliable adoption of data-driven models in real-world applications. This paper suggests that leveraging existing knowledge of structure-rich spaces from classical mechanics and differential geometry remains an underutilized opportunity for developing data-driven models that can generalize at a structural level. The presented method for `nudging' an existing physics-based model towards observation data achieved good generalization, even with a simple data-driven model, by preserving structural symmetry and demonstrates how such knowledge may be exploited to develop more parsimonious data-driven models. There are a wide range of potential applications of this method in digital twinning, where near real-time assimilation of live data is often required, while at the same time ensuring that models can generalize across operating conditions.




\FloatBarrier 
\balance
\bibliography{main}
\bibliographystyle{icml2025}
\appendix
\onecolumn
\section{Material properties for high-fidelity simulation}
\begin{table*}[h]
\begin{center}
\begin{tabular}{llll}
\toprule
Genre & Property & Target (measurements) & Misspecified (physics)\\
\midrule
Physical properties & volume & 1.8 $m^3$ & 3.6 $m^3$ \\
                    & layer thickness & 0.2 $m$ & 0.4 $m$ \\
Thermophysical material properties & conductivity & 0.72 $W/mK$ & 0.2 $W/mK$\\
                                & density & 1920 $kg/m^3$ & 1920 $kg/m^3$\\
                                & specific heat capacity & 780 $J/kgK$ & 780 $J/kgK$ \\
Convection coefficients & outdoor convection & 25  $W/m^2K$ & 20  $W/m^2K$ \\
Theremophysical air properties & air density & 1.2 $kg/m^3$ & 1.2 $kg/m^3$ \\
                                & air specific heat capacity & 100 $J/kgK$ & 100 $J/kgK$ \\
                               
\bottomrule
\end{tabular}
\label{table:simulation}
\caption{Physical and thermodynamic properties used to generate target data via numerical analysis in EnergyPlus and to initialize the state space model optimisation, respectively.}
\end{center}
\end{table*}

\section{Sweep over Random Forest and XGBoost forest size}
\begin{figure}[ht] 
\vskip 0.2in
\begin{center}
\centerline{\includegraphics[width=0.6\columnwidth]{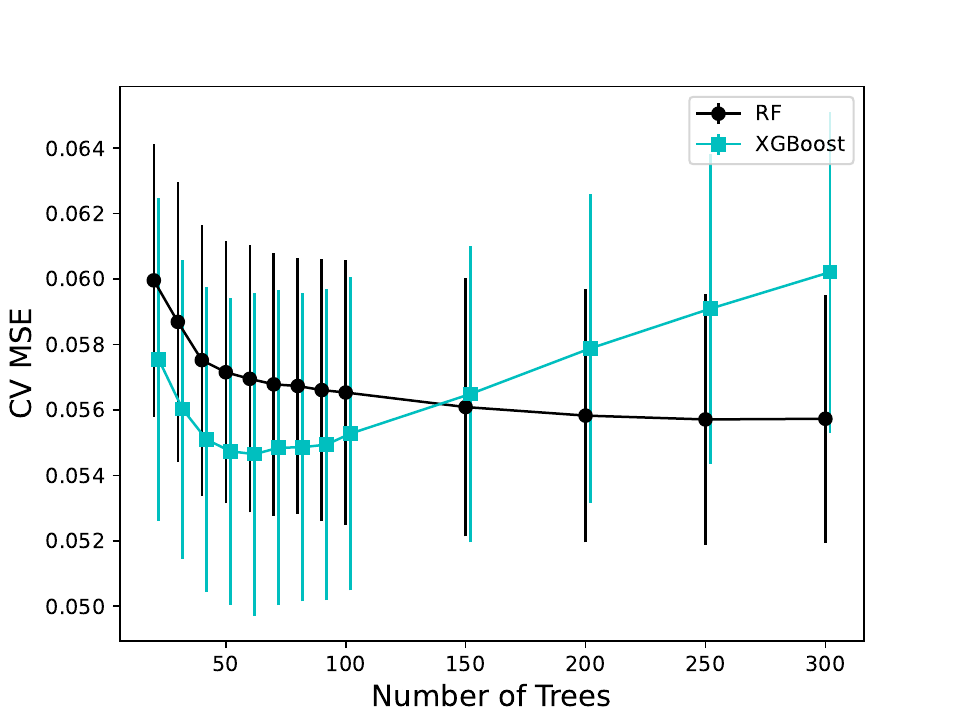}}
\caption{Five-fold cross validated MSE, for a sweep of forest sizes.}
\label{fig:rf_sweep}
\end{center}
\vskip -0.2in
\end{figure}
Figure \ref{fig:rf_sweep} displays the five-fold cross validated (CV) MSE, for a sweep of forest sizes. The training portion of the London dataset was used to perform this sweep, with a forest size of 250 trees selected for the random forest and 60 for XGBoost. 

\FloatBarrier
\section{Convergence of optimized state space models}\label{sec:app_lssm}

\begin{figure}[h] 
\begin{center}
\centerline{\includegraphics[width=0.6\columnwidth]{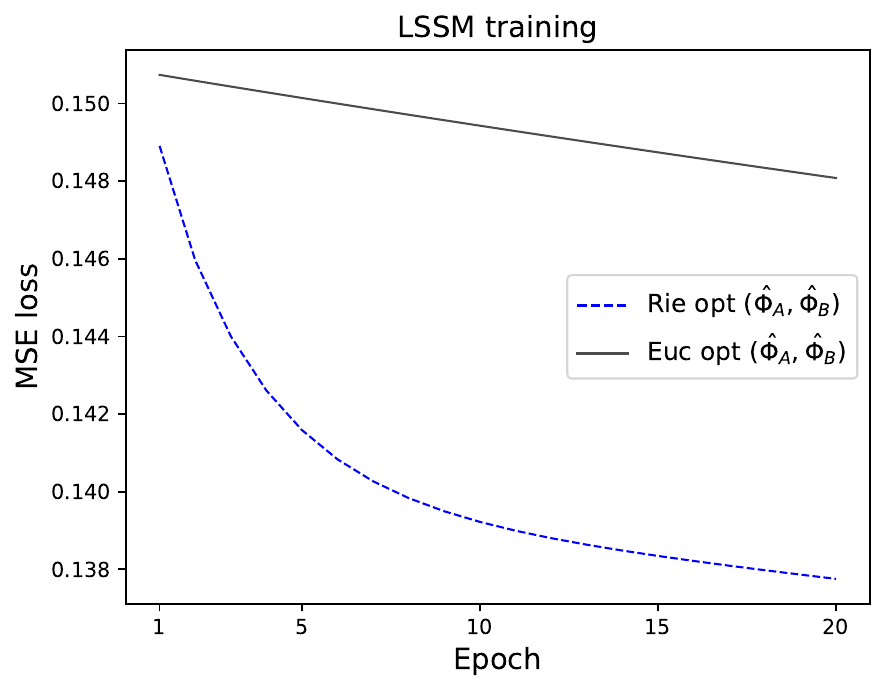}}
\caption{MSE loss per epoch for optimized LSSMs.}
\label{fig:lssm_train}
\end{center}
\end{figure}

\begin{figure}[h] 
\vskip 0.2in
\begin{center}
\centerline{\includegraphics[width=0.6\columnwidth]{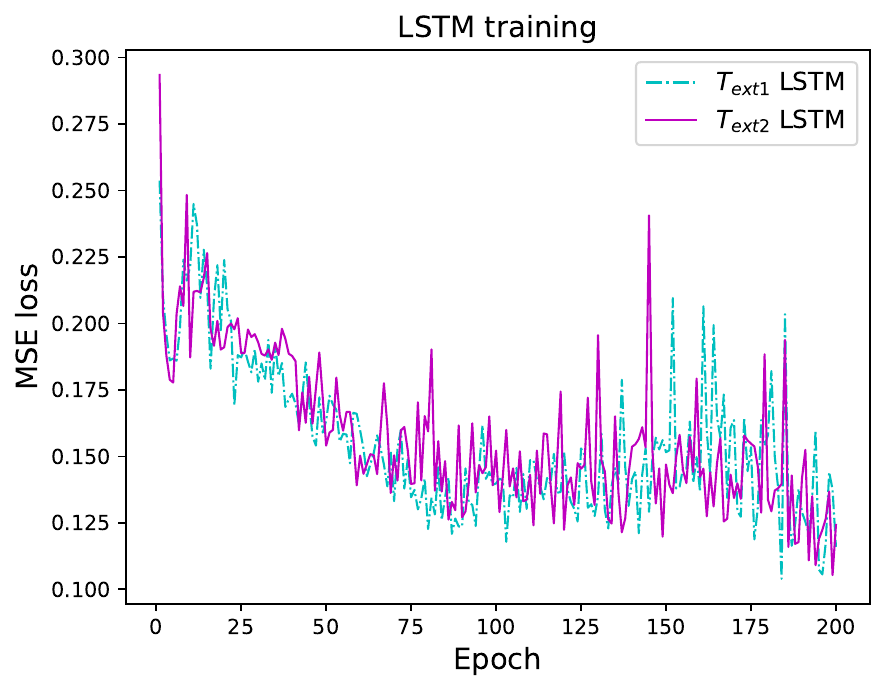}}
\caption{MSE loss per epoch for the LSTMs trained on $T_{ext1}$ and $T_{ext2}$.}
\label{fig:lstm_train}
\end{center}
\vskip -0.2in
\end{figure}

Figure \ref{fig:lssm_train} displays the MSE loss per epoch of the model-based approaches $\mathbf{Rie \text{ } opt}$ and $\mathbf{Euc \text{ } opt}$ during training on the London dataset. Note that $\mathbf{Rie \text{ } opt}$, trained by optimizing on the Riemannian manifold, converged significantly faster than the model optimized in Euclidean space. 

Figure \ref{fig:lstm_train} displays the convergence of the investigated LSTM architectures in training. Considerable instability was observed in LSTM training and testing on the London dataset, with best results achieved by learning $T_{ext1}$ and $T_{ext2}$ independently. Both LSTMs used 64 hidden layers, with a window size of 100. There is a strong seasonal variation to both the London and Chicago dataset. The poor performance of the LSTMs was attributed to the relatively small size of the training dataset, which limited the window size and made it difficult for the LSTMs to capture seasonal variations. 

\FloatBarrier

\section{Plots displaying $T_{ext2}$}
\begin{figure}[h] 
\begin{center}
\centerline{\includegraphics[width=1.0\columnwidth]{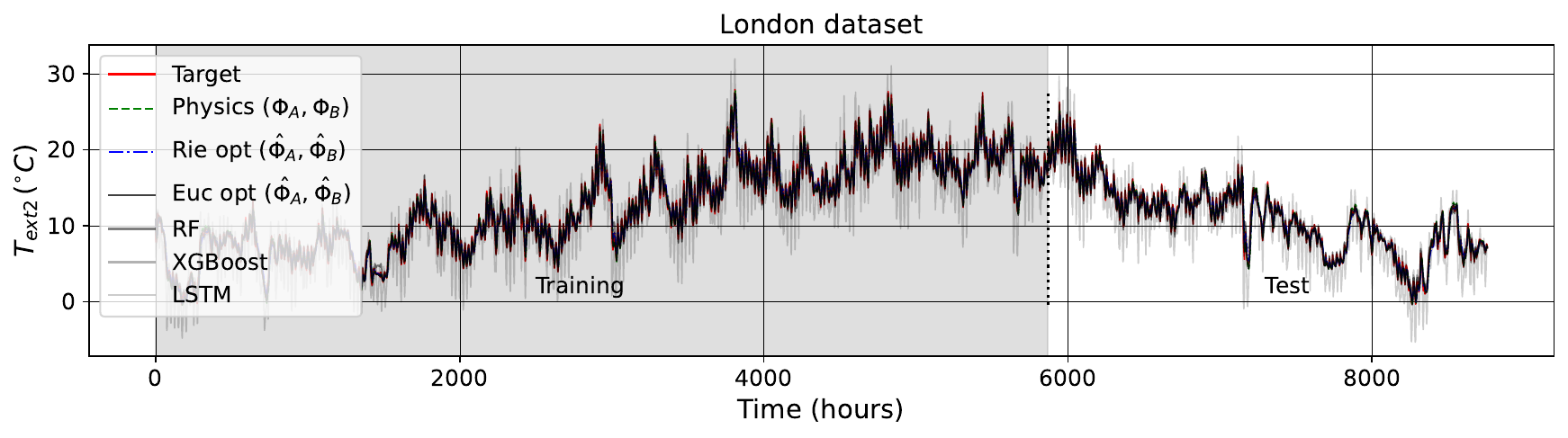}}
\caption{Model training and testing on $T_{ext2}$ for the London dataset.}
\end{center}
\label{fig:london_tex2}
\end{figure}

\begin{figure}[h] 
\begin{center}
\centerline{\includegraphics[width=1.0\columnwidth]{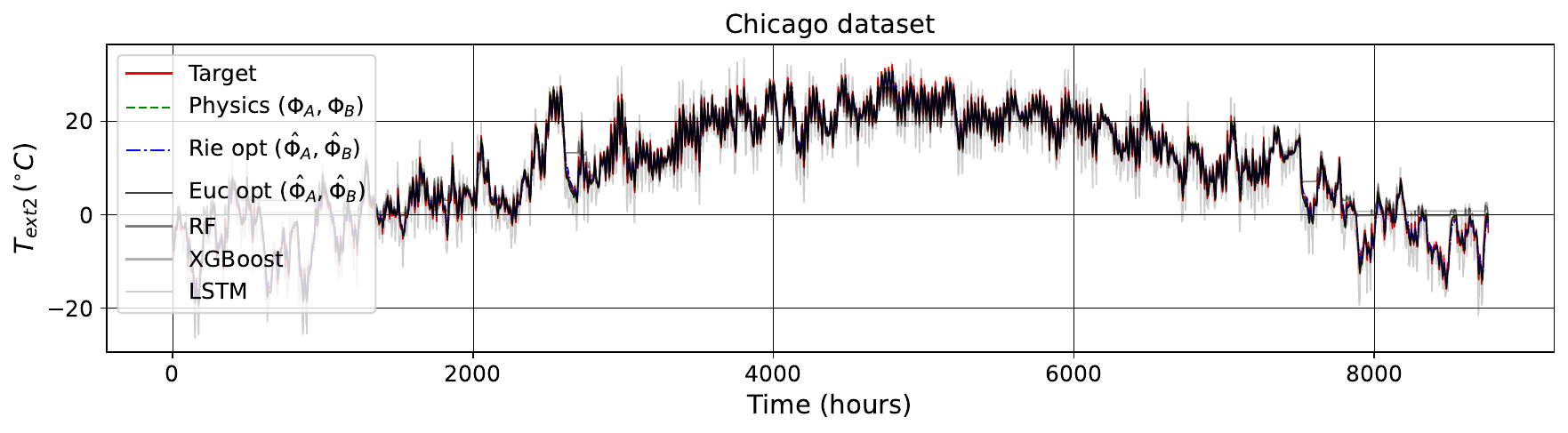}}
\caption{Model testing on $T_{ext2}$ for the Chicago dataset.}
\label{fig:chicago_tex2}
\end{center}
\end{figure}

The exterior surface temperature, $T_{ext2}$, is highly correlated with the external forcing acting on the system, $U_t$. As can be seen from Figures~{\ref{fig:london_tex2}} and \ref{fig:chicago_tex2} all of the investigated models predicted this variable accurately. For that reason, emphasis was placed on accurate estimation of $T_{ext1}$ in the discussion in section~{\ref{sec:results}}.

\end{document}